\documentclass[runningheads]{llncs}

\usepackage[T1]{fontenc}
\usepackage{graphicx}
\usepackage{amsfonts}
\usepackage{amsmath}
\usepackage{amssymb}
\usepackage{physics}
\usepackage{svg}
\usepackage{pdfpages}
\usepackage{siunitx}
\begin{document}
\title{Parkinson's Disease Detection from Resting State EEG using Multi-Head Graph Structure Learning with Gradient Weighted Graph Attention Explanations}
\titlerunning{Multi-Head GSL for Parkinson's Detection}
\author{Christopher Neves\inst{1}\orcidID{0009-0006-9329-6673} \and
Yong Zeng\inst{2}\orcidID{0000-0001-6678-271X} \and
Yiming Xiao\inst{3}\orcidID{0000-0002-0962-3525}}

%index{Neves, Christopher}
%index{Zeng, Yong}
%index{Xiao, Yiming}

\authorrunning{C. Neves et al.}
\institute{Department of Computer Science and Software Engineering, Concordia University, Montreal, QC, Canada \email{christopher.neves@hotmail.ca} 
\and
Concordia Institute for Information Systems Engineering, Concordia University, Montreal, QC, Canada \email{yong.zeng@concordia.ca}\\
\and
Department of Computer Science and Software Engineering, Concordia University, Montreal, QC, Canada \email{yiming.xiao@concordia.ca}}

% Camera ready checklist
% - Sign copyright form
% - Submission must include all latex source files (pictures, bib files, etc)
% - Submission must also include pdf of paper
% - Yong Zeng ORCID

\maketitle              % typeset the header of the contribution
\begin{abstract}
Parkinson’s disease (PD) is a debilitating neurodegenerative disease that has severe impacts on an individual’s quality of life. Compared with structural and functional MRI-based biomarkers for the disease, electroencephalography (EEG) can provide more accessible alternatives for clinical insights. While deep learning (DL) techniques have provided excellent outcomes, many techniques fail to model spatial information and dynamic brain connectivity, and face challenges in robust feature learning, limited data sizes, and poor explainability. To address these issues, we proposed a novel graph neural network (GNN) technique for explainable PD detection using resting state EEG. Specifically, we employ structured global convolutions with contrastive learning to better model complex features with limited data, a novel multi-head graph structure learner to capture the non-Euclidean structure of EEG data, and a head-wise gradient-weighted graph attention explainer to offer neural connectivity insights. We developed and evaluated our method using the UC San Diego Parkinson's disease EEG dataset, and achieved 69.40\% detection accuracy in subject-wise leave-one-out cross-validation while generating intuitive explanations for the learnt graph topology.

\end{abstract}
\section{Introduction}

Parkinson's Disease (PD) is the second most common neurodegenerative disorder worldwide \cite{tolosa_challenges_2021}. Primarily characterized by motor symptoms, the complex disease can also include psychiatric and cognitive issues. MRI-based biomarkers have attracted major attention, including biochemical alteration shown in quantitative MRI and structural/functional connectivity changes revealed by diffusion and functional MRI \cite{greenspan_explainable_2023}. However, electroencephalography (EEG), which records electric signals from a network of locations on the scalp is a much more cost-effective neuroimaging tool with higher temporal resolution than MRI that has also been investigated to provide neurological insights and potential biomarkers for the disease. This is especially true for remote or less privileged regions, where MRI scanners are difficult to access.

Recently, deep learning (DL)-based techniques have provided excellent outcomes for EEG analysis, but several challenges remain. \textbf{First}, most existing DL techniques for EEG rely on Convolutional Neural Networks (CNNs) that aggregate signals across channels \cite{dose_end--end_2018}\cite{lawhern_eegnet_2018}, but such approaches can miss key spatial characteristics of EEG signals, limiting clinically relevant brain connectivity insights and explainability. \textbf{Second}, to better incorporate spatial information, graph neural networks (GNNs) that model different EEG sensors and their relationships as nodes and edges of a graph (often represented as an adjacency matrix) have been proposed. However, although stationary connectivity metrics, such as the Pearson Correlation Coefficient (PCC) or Absolute Cross-Correlation (ACC) are straightforward for deriving the graph for GNN, they often fail to capture non-stationary connectivity, overestimate the correlation between adjacent nodes due to mixing of electrical signals over the scalp surface, and may not provide true functional connectivity insights in many situations. \textbf{Third}, EEG data sampled at high frequencies often involves very long sequences, which can pose challenges for commonly used sequential DL models to capture task-relevant features. Recently, Li \textit{et al.} \cite{li_what_2022} tackled this issue with an effective convolutional model called Structured Global Convolution (SGConv) that has surpassed state-of-the-art sequence models, including Transformers \cite{vaswani_attention_2017} and Structured State Spaces \cite{gu_efficiently_2022}, by designing a global convolutional kernel that can span the length of the entire sequence. \textbf{Finally}, compared with other medical imaging data, the typically small cohort sizes of EEG datasets can pose challenges for developing robust DL techniques in the domain. 

In this work, we aim to address the aforementioned issues with three contributions. \textbf{First}, we combined structured global convolutions \cite{li_what_2022} and self-supervised contrastive learning to better model complex and long sequences of EEG data with a limited cohort for the first time; \textbf{Second}, we proposed a novel dynamic multi-head graph structure learning technique to learn the adjacency matrix of the underlying EEG data without imposing potential biases in contrast to conventional static GNN methods; \textbf{Third}, to enhance explainability of our DL model for potential clinical insights, we introduced a new technique based on head-wise gradient-weighted attentions to generate an informative adjacency matrix to reveal key task-relevant connectivities in the learnt graph. The proposed method is demonstrated for PD detection with resting state EEG.

\section{Related Works}

To date, several GNN-based methods \cite{klepl_graph_2023} have been explored for EEG analysis, particularly for seizure detection in epilepsy. Traditionally, manually defined EEG features, such as Short Time Fourier Transform \cite{covert_temporal_2019}, power spectral density \cite{jin_eeg-based_2021}, and selective frequency bands \cite{song_eeg_2020} have been used in machine/deep learning, but can introduce biases while being time-consuming and expertise-demanding. Therefore, automatic feature extraction methods have become more desirable to reduce biases and improve efficiency. Among these, Dissanayake \textit{et al.} \cite{dissanayake_geometric_2022} and Sun\textit{ et al.} \cite{sun_dual-branch_2022}  used stacked Long Short-Term Memory (LSTM) networks and Transformers to generate feature embeddings. Li \textit{et al}. \cite{li_what_2022} proposed the Structural Global Convolution, which showed superior ability to model long and complex sequential signals than prior approaches. Using EEG feature embeddings as node features, different GNN designs incorporating temporal features and spatial properties of EEG data have been devised. One notable trend is the rise of attention-based GNNs, which allow for the visualization of salient edges relevant to the designated tasks to enhance DL model transparency. He \textit{et al.} \cite{he_spatialtemporal_2022} used a graph attention network (GAT) in conjunction with a bi-directional LSTM for seizure detection and Demir \textit{et al.} \cite{demir_eeg-gat_2022} used a GAT with additional temporal convolutions to decode motor signals. To mitigate issues with static graphs, Tang \textit{et al.} \cite{pmlr-v209-tang23a} and Song \textit{et al.} \cite{song_eeg_2020} employed the concept of attention to learn the graph adjacency matrix instead of the attention weights between nodes (as in GATs). However, both of their formulations use a single attention head. In EEG-based PD analysis, Chang \textit{et al}. \cite{chang_eeg-based_2023} developed a GNN that learns attention coefficients with a graph sparsity constraint to modulate the node feature vectors for PD detection during an auditory oddball task. Further explorations are still required to enhance the efficiency, accuracy, robustness, and transparency of DL-based EEG analysis, especially for GNN-based approaches.

\section{Methods and Materials}

Figure \ref{fig:arch} outlines an overview our proposed DL architecture, which is composed of a feature encoder (LongConv feature encoder), a multi-head graph structure learner (MH-GSL), a Chebyshev GNN, and a classifier made of fully connected layers for PD vs. Healthy classification.

\subsection{Feature Encoder with Contrastive Learning}

Following the success of Structured Global Convolutions (SGConv) \cite{li_what_2022} for modeling long sequential data in deep learning tasks, we incorporate it into our EEG feature encoder design, which encodes the input EEG signal to $\Tilde{X}_e \in \mathbb{R}^{C \times d_m}$ ($C$ is the number of channels and $d_m$ is model dimension). Specifically, we follow the feature extraction network setup in the work of Vetter \textit{et al.} \cite{vetter_generating_2023}, who modify the Structured Global Convolution layer from its original formulation to have more fine-grained control over its kernel size (referred to as SLConv in Fig. \ref{fig:arch}). The feature extraction network (called LongConv) consists of interleaved masked 1D convolutions, which project the input channels to a set of hidden ones while SLConv layers extract long-range temporal information from each hidden channel. Each masked 1D convolution is followed by a batch normalization layer and a GELU activation. In our adapted LongConv feature encoder design, we add an additional max pooling operation followed by a 1D convolution (Conv1D) to their network structure before the MH-GSL and Chebyshev GNN layers. To alleviate some of the issues presented by the large inter-subject variability of EEG and the relatively small dataset size, we pretrained the LongConv encoder using the SimCLR \cite{chen_simple_2020} framework. First proposed for natural images, SimCLR learns self-supervised data representation by maximizing agreement between differently augmented versions of the same data sample based on a contrastive loss in the latent space. For EEG contrastive learning (CL), we adopted the data augmentations by Mohsenvand \textit{et al}.\cite{pmlr-v136-mohsenvand20a}, including combinations of random additive Gaussian noise, random signal masking, a flip along either the signal or electrode dimension or random DC shifts. During training, we used a simple two-layer feed forward network as the projector after the LongConv encoder to obtain a latent space representation used to compute the InfoNCE loss \cite{oord_representation_2019}. We used a learning rate of 0.0001, a temperature of 0.005 \cite{pmlr-v136-mohsenvand20a}, and a batch size of 100 over 160 SimCLR training epochs.

\subsection{Multi-Head Graph Structure Learner}

Graph topology of EEG signals obtained from stationary connectivity measures and/or the physical distance between electrodes for GNN learning can be misleading and sub-optimal. To tackle this, we proposed a novel graph structure learner (GSL) using multi-head attention. Based on the graph structure layer by Tang \textit{et al.} \cite{pmlr-v209-tang23a}, which adopts the self-attention mechanism \cite{vaswani_attention_2017} to learn edge weights, we extended this approach to include multiple attention heads. Thus, the resulting graph structure learner can attend to different graph representations (adjacency matrices) in parallel, with each attention head providing the edge weights for its paired graph representation. Then, each head-wise learnt graph representation, together with the encoded EEG features are passed to a Chebyshev GNN, updating the features with the learnt spatial relationships. The output of the Chebyshev GNN for each head is then concatenated and projected back to the model dimension $d_m$ using a linear layer. The adjacency matrix $A_h \in \mathbb{R}^{C \times C}$ for a single attention head $h$ out of $H$ heads is given by:

\begin{equation}
    \begin{split}
        Q_h = \Tilde{X}_e W_{q_h}, K_h = \Tilde{X}_e W_{k_h} \\
        A_h = softmax(\frac{Q_h K_h^T}{\sqrt{d_K}})
    \end{split}
\end{equation}

\noindent where $\Tilde{X}_e \in \mathbb{R}^{C \times d_m}$ are the feature embeddings, and $W_{q_h}$ and $W_{k_h}$ are the parameter matrices projecting $\Tilde{X}_e$ to query $Q_h$ and key $K_h$, respectively. 

\subsection{Graph-based EEG Classification}

\begin{figure}
    \centering
    \includegraphics[trim=0cm 23cm 0cm 3cm, clip=false, width=\textwidth]{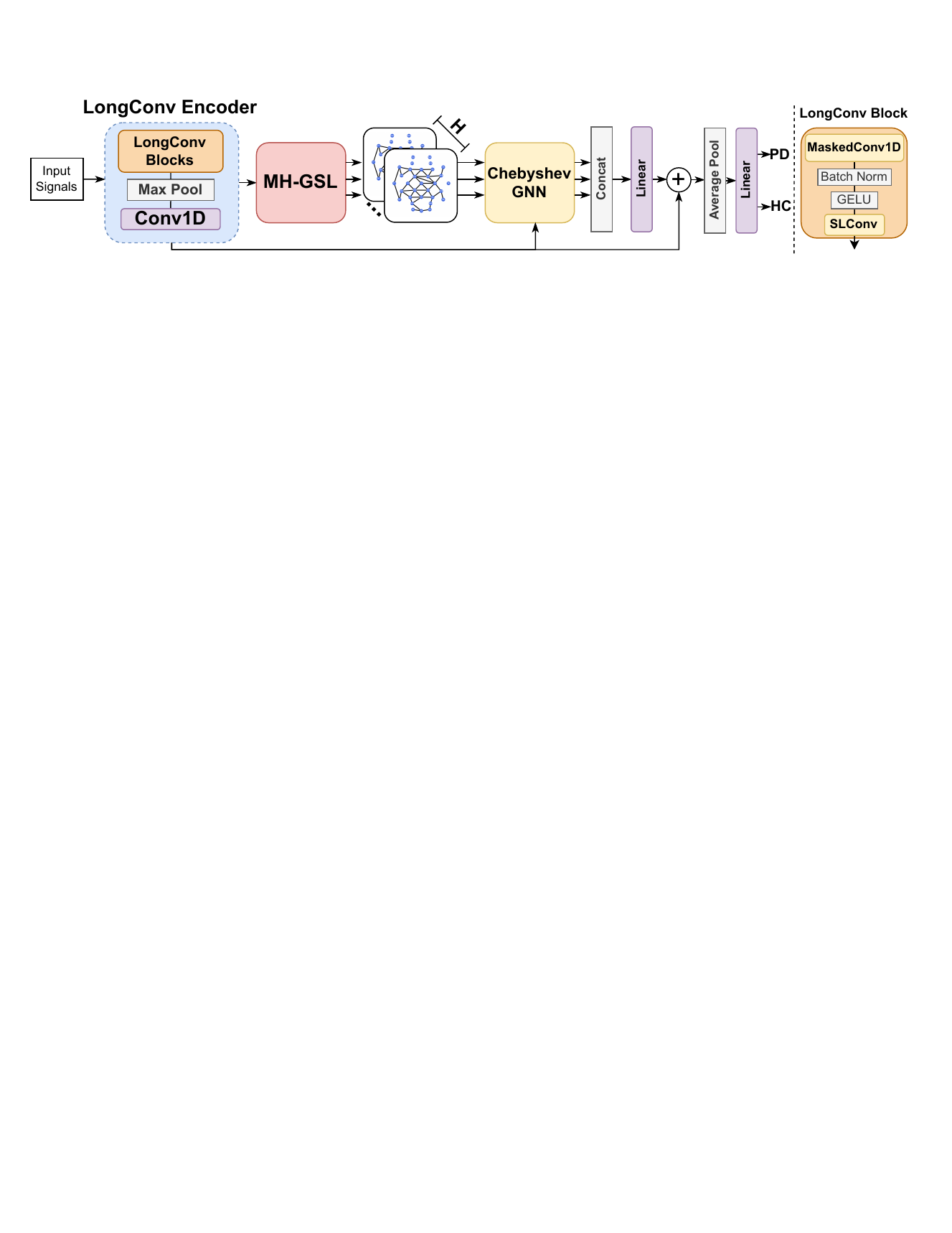}
    \caption{Overview of the model architecture for PD detection.}
    \label{fig:arch}
\end{figure}

As shown in Fig. \ref{fig:arch}, the final EEG classification is achieved by first adding the head-wise aggregated output from the Chebyshev GNN and EEG feature embeddings from the temporal feature encoder, and average pooling the result along the electrode dimension to yield a final representation of shape $\Tilde{X}_g \in \mathbb{R}^{C \times 1}$. A linear layer is then used to perform Healthy vs. PD classification. We use the cross-entropy loss and AdamW optimizer \cite{loshchilov_decoupled_2019} to train our model. Here, we use the Chebyshev GNN in our model, as it has previously been used for EEG analysis \cite{dissanayake_geometric_2022} \cite{loshchilov_decoupled_2019} and is an effective method of integrating an adjacency matrix with EEG feature embeddings by efficiently approximating graph convolutions using Chebyshev polynomials.

\subsection{Head-wise Gradient-Weighted Graph Attention Explainer}

In multi-head self-attention networks, the average or maximum of the head-wise attention scores \cite{velickovic_graph_2017} are often used to provide graph explainations, but this could be insufficient as some heads may carry greater contributions for decision-making. Inspired by the work of Rasoulian \textit{et al}. \cite{rasoulian_weakly_2023}, where head-wise gradient-weighted self-attention maps were used to improve the specificity of the attention map, we adapted the core idea for GNN-based EEG analysis. Specifically, we obtain a graph explanation by first weighing the head-wise graph representation $A_h$ with the norm of its gradient based on the class activation. Then, the final adjacency matrix $A \in \mathbb{R}^{ C \times C}$ is generated as:

\begin{equation}
    A = \frac{1}{H} \sum_{h=1}^H \norm{\frac{\partial Y}{\partial A_h}} \cdot A_h
\end{equation}

\noindent where $H$ is the number of attention heads and $Y$ is the target class to generate a graph representation for. Finally, $A$ is  thresholded to keep the attention scores within two standard deviations from the mean, and then are normalized to [0,1].

\subsection{Dataset and Preprocessing}
We used the UC San Diego Parkinson's disease resting-state EEG (rs-EEG) dataset \cite{san_diego_pd} for our study. The dataset contains the resting-state data of 15 PD patients (63.2$\pm 8.2$ years, 8 females) and 16 healthy controls (63.5$\pm 9.6$ years, 9 females). All PD patients had mild to moderate disease severity. Each participant had at least 3 minutes of resting state data recorded using a 32-channel Biosemi ActiveTwo EEG system (sampling rate = 512 Hz). We minimally preprocessed each subject's EEG by first setting the reference to the mean of the EXG7 and EXG8 mastoid electrodes and band-pass filtered the raw signal to 0.5-80 Hz. The data was then segmented into 2 sec of non-overlapping windows, resulting in 90 trials per participant.

\subsection{Experimental Setup and Ablation Studies}

To assess the classification performance of our proposed framework, we compared it against a variety of DL models and configurations. With CNN methods dominating EEG analysis, as a baseline, we re-implemented the method by Dose \textit{et al.} \cite{dose_end--end_2018} that showed great success on small datasets. To further validate the benefits of each design component of our method, we performed a series of ablation studies. \textbf{First}, to confirm the contribution of the Chebyshev GNN, we compared the full version of our method (CL-Encoder+Freeze) against PD detection only based on the temporal feature encoder (LongConv Encoder). \textbf{Second}, to verify whether our multi-head GSL had a positive impact on the network performance, we replaced the learnt graph structure input to the Chebyshev GNN with a static graph based on PCC, and evaluate the classification accuracy against the original design (``Full Model w/o MH-GSL vs. Full Model with MH-GSL", both without CL). \textbf{Third}, to quantify the performance gain from the SimCLR framework, we compared the proposed frameworks with and without self-supervised pre-training (``CL-Encoder+Freeze  vs. Full Model with MH-GSL"). \textbf{Finally}, as some studies demonstrated the benefit of finetuning pre-trained feature encoder, we further tested our proposed method by finetuning the feature encoder weights that were pre-trained using the SimCLR framework, and compared the outcome to freezing the feature encoder weights after SimCLR pre-training (``CL-Encoder+Finetune vs. CL-Encoder+Freeze"). We computed classification accuracy, precision and recall, macro F1-score, and AUC metrics for all experimental setups over 3 random seeds (i.e., model weight initialization). 

We trained and evaluated all configurations using a leave-one-out cross-validation, where a single subject was used for testing and the rest for training to avoid data leakage. For each fold, two subjects (one healthy and one PD) were randomly selected from the training data as a validation set. Unlike the more common sample-wise cross-validation in EEG-related DL algorithms, our subject-wise strategy can better assess the generalizability of the proposed framework to unseen subjects. Each model was trained with a batch size of 8 with a MultiStep learning rate (LR) scheduler at an initial LR of 1E-4 and a gamma of 0.1. The MH-GSL model was trained using 2 attention heads and the Chebyshev GNN used a single layer with K=5 and a dropout rate of 0.2.

\section{Results}

\begin{table}
\centering
\addtolength{\leftskip} {-2cm} 
\addtolength{\rightskip}{-2cm}
\caption{PD vs. Healthy classification performance for all model configurations.}\label{tab1}
\begin{tabular}{l|c|c|c|c|c}
\hline
Method &  Accuracy \% &AUC & F1-Score & Precision & Recall\\
\hline
\hline
LongConv Encoder &  64.68$\pm$1.85 & 0.638$\pm$0.039 & 0.643$\pm$0.017 &0.649$\pm$0.020& 0.644$\pm$0.018\\
Full Model w/o MH-GSL &  66.97$\pm$1.29  &0.670$\pm$0.013&0.663$\pm$0.009& 0.677$\pm$0.021&0.666$\pm$0.011\\
Full Model with MH-GSL & 67.73$\pm$0.85 & \textbf{0.715$\pm$0.024} & 0.672$\pm$0.009 & 0.682$\pm$0.009 & 0.674$\pm$0.009\\
\hline
CL-Encoder + Freeze& \textbf{69.40$\pm$1.59} & 0.656$\pm$0.036 & \textbf{0.682$\pm$0.016} &  \textbf{0.716$\pm$0.021} & \textbf{0.688$\pm$0.015}\\
CL-Encoder + Finetune& 66.34$\pm$2.68 & 0.707$\pm$0.010 & 0.658$\pm$0.030 & 0.668$\pm$0.026 & 0.660$\pm$0.027\\
\hline
CNN classifier \cite{dose_end--end_2018}&  62.99$\pm$4.07 & 0.640$\pm$0.061 & 0.629$\pm$0.040 & 0.629$\pm$0.041 &0.629$\pm$0.040 \\
\hline
\end{tabular}
\end{table}

\begin{figure}
    \centering
    \includegraphics[trim=1cm 5cm 1cm 1cm, clip=false, width=0.7\textwidth]{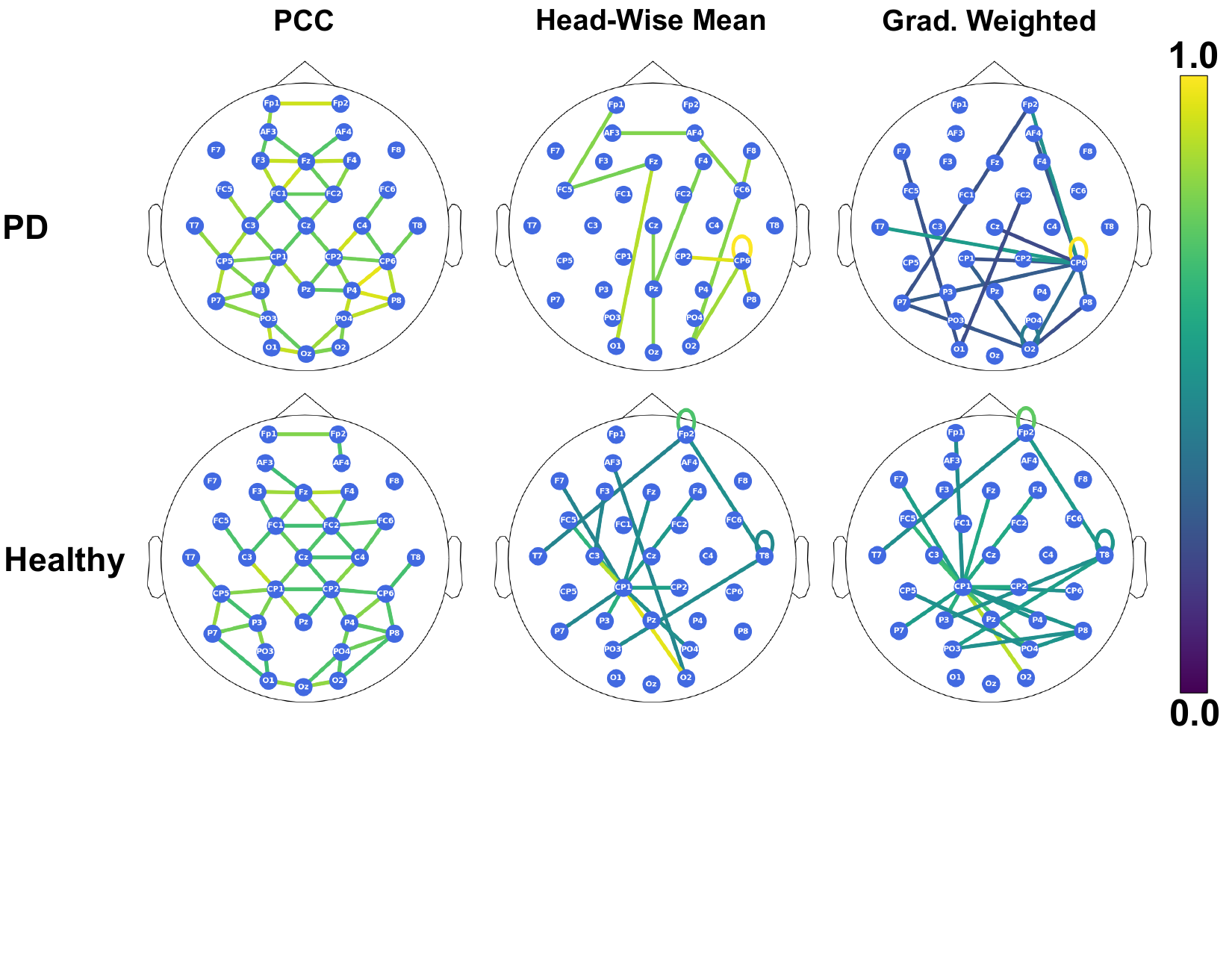}
    \caption{Group-wise mean adjacency matrices for PD and healthy subjects for static PCC, mean head-wise attention, and gradient-weighted mean head-wise attention.}
    \label{fig:adj_mat}
\end{figure}

\noindent We present the PD vs. Healthy classification performance of all experiments in Table \ref{tab1}, and with an accuracy of 69.40±1.59\%, our proposed method (CL-Encoder+Freeze) outperformed the CNN baseline \cite{dose_end--end_2018} (accuracy=62.99±4.07\%) and the other model configurations. For the ablation studies, we confirmed the positive impact of Chebyshev GNN, multi-head graph structure learner, simCLR-based encoder pretraining. Furthermore, between CL-Encoder+Finetune and CL-Encoder+Freeze, further finetuning the feature encoder during full model training decreased all evaluation metrics by 3$\sim$5\%. In addition, Fig. \ref{fig:adj_mat} presents the resulting adjacency matrices averaged for the PD and HC groups for all correctly classified samples based on static PCC-based graphs, mean of head-wise attentions from our MH-GSL, and gradient-weighted mean head-wise attention also from our MH-GSL. The gradient-weighted adjacency matrices show a greater amount of connections towards the inion (back) of the skull compared to their non-weighted counterparts. The PCC graphs show almost exclusively connections between neighboring nodes.

\section{Discussion}

Our novel multi-head graph structure learner presents a more dynamic approach that establishes task-driven graphs with improved performance in comparison to static connectivity graphs. This observation agrees with previous studies \cite{pmlr-v209-tang23a}. So far, despite many attempts to learn graph edge weights using attention mechanisms\cite{pmlr-v209-tang23a}\cite{chang_eeg-based_2023}, very few extended their formulation to include multiple attention heads despite their great success in vision and language tasks. Different from approaches where attention scores are multiplied with the features in an initial graph \cite{chang_eeg-based_2023}\cite{li_dynamical_2024}, we directly learn different node features for each adjacency matrix from MH-GSL in parallel, and finally concatenate them for classification. After testing different numbers of attention heads (2, 4 and 8), we found that two heads yielded superior performance for this task. To the best of our knowledge, we are the first to propose a head-wise gradient weighted graph attention explanation to obtain visual interpretation for task-relevant brain connectivity properties. This approach helps further highlight task-relevant graph information. Figure \ref{fig:adj_mat} reveals that graphs learnt  with our method focus more on global connections across the scalp, and overcome the overemphasis on adjacent connections seen in commonly used stationary graphs. It is also interesting to note that weighing the head-wise adjacency matrices by the norm of their gradients results in a more connected graph structure compared to its unweighted counterpart. Qualitatively, the number of connections seems to greatly increase with gradient-weighing for PD subjects, thus showing a higher connection count to be important for classification. Although an increase in functional connectivity has been shown in PD patients in resting state EEG studies \cite{bosch_functional_2022}, additional analysis of the generated edge explanations is required before drawing neuroscientific conclusions. Nevertheless, the presented technique offers great potential for deriving important connectivity information for the disorder under study. We will further validate the physiological significance of the resulting graph explanation with joint EEG-fMRI studies as the relevant insights could be of more value than PD vs HC classification. 

In our experiments, we adopted a subject-wise leave-one-out cross-validation instead of a sample-wise one seen in many reports. The latter approach is often used to accommodate limited subjects in EEG datasets, but can easily cause data leakage issues, resulting in exaggerated accuracy. When adopting this commonly used strategy, our model yields near perfect classification results ($\sim$98\% accuracy) potentially due to memorizing subject-specific details instead of task relevant ones. To help address limited data size, we employed contrastive learning to enhance the robustness of our feature encoder, and its benefit is evident in our experiment (1.67\% accuracy increase). In comparison to fMRI and task-based EEG, rs-EEG is easier to acquire, but requires more sophisticated feature extraction techniques. Through PD detection, we demonstrated great performance of the proposed DL method and a novel graph explanation technique. We will showcase its adaptability in extended applications in the future.

\section{Conclusion}
We have developed a novel GNN technique for PD detection from resting state EEG based on dynamic graph structure learning, with a head-wise gradient-weighted graph explainer. In addition, we demonstrated the benefit of contrastive learning in efficient and robust feature extraction from a small cohort. With thorough evaluations and ablation studies, the performance of our proposed method has a great potential to offer clinical insights for PD and extended neurological applications with more accessible EEG sensors.  

\begin{credits}
\subsubsection{\ackname} We acknowledge the support of the Natural Sciences and Engineering Research Council of Canada and Y. Xiao is supported by the Fond de la Recherche en Santé du Québec (FRQS—chercheur boursier Junior 1).

\subsubsection{\discintname}
The authors have no competing interests to declare
that are relevant to the content of this article.
\end{credits}

\bibliographystyle{splncs04}
\bibliography{mybibliography}
\end{document}